# Cyberbullying Detection Using Deep Neural Network from Social Media Comments in Bangla Language

Md Faisal Ahmed , Zalish Mahmud, Zarin Tasnim Biash, Ahmed Ann Noor Ryen, Arman Hossain, Faisal Bin Ashraf

Department of Computer Science and Engineering, Brac University,
Dhaka, 1212, Bangladesh

*Abstract* - Cyberbullying or Online harassment detection on social media for various major languages is currently being given a good amount of focus by researchers worldwide. Being the seventh most speaking language in the world and increasing usage of online platform among the Bengali speaking people urge to find effective detection technique to handle the online harassment. In this paper, we have proposed binary and multiclass classification model using hybrid neural network for bully expression detection in Bengali language. We have used 44,001 users' comments from popular public Facebook pages, which fall into five classes - Non-bully, Sexual, Threat, Troll and Religious. We have examined the performance of our proposed models from different perspective. Our binary classification model gives 87.91% accuracy, whereas introducing ensemble technique after neural network for multiclass classification, we got 85% accuracy.

I. INTRODUCTION

Natural Language Processing (NLP) is the innovation used to aid computers to comprehend the human's natural language. NLP is a branch of artificial intelligence that deals with the interaction between computers and humans utilizing the natural language. A definitive goal of NLP is to read, decode, comprehend, and make sense of the human dialects in a way that is significant. There are generally five steps to process natural language - Lexical Analysis, Syntactic Analysis, Semantic Analysis, Discourse Integration, and Pragmatic Analysis [3]. Lexical Analysis involves identifying and analyzing the structure of words. Lexicon of a language implies the assortment of words and phrases in a language. Lexical analysis divides the entire group of texts into paragraphs, sentences, and words. Syntactic Analysis (Parsing) involves analyzing the words present in the sentence for grammar and organizing the words in a way that shows the relationship among the words. For example, the sentence "I go to by school" is dismissed by the English syntactic analyzer. Similarly, the sentence "বর্ষার রোদে প্লাবনের সৃষ্টি হয়" is rejected by the Bengali syntactic analyzer. In semantic analysis, the dictionary meaning of the word is being drawn from the text. In other words, the text is checked for meaningfulness. This is done with the help of mapping syntactic structures and objects in the task domain. The semantic analyzer disregards sentences such as "hot ice-cream" in English or "আঙ্গুল ফুলে কলাগাছ" in Bengali. Discourse Integration involves the meaning of a sentence that relies on the significance of the sentence just before it. Furthermore, it also realizes the meaning of the immediately succeeding sentence. Lastly, in the pragmatic analysis step, information exchanged is re-deciphered on what it really implied. It includes determining those parts of language which require true information or real-world knowledge. Natural Language Processing allows computers to understand text, hear speech, decipher it, measure sentiment and figure out which parts are important [4]. NLP has very important usage for language translation applications such as Google Translate [2]. Word Processors such as Microsoft Word and Grammarly use NLP to check grammatical accuracy of texts [2]. Similarly, we have used NLP to analyze the sentiment of the Bengali text. This allows us to differentiate whether the sentence is a bully expression or not.

Cyberbullying or online harassment is the utilization of electronic correspondence to menace an individual, ordinarily by sending messages of an intimidating or compromising nature. In the era of social media and online networking, the usage of offensive and aggressive words has increased significantly. These comments build up a culture of disrespect in cyberspace [5]. In earlier years, cyberbullying was not properly paid attention to and was overlooked. The explanation was low participation of users in the social networking platform and it was recommended to screen off or detach in the event when one would get harassing remarks. Be that as it may, presently the situation is completely changed. A 2019 study shows that out of every 100 women being cyber harassed, 70 are aged 15-25 years. Among the harassment allegations and cases that come to the country's only cybercrime tribunal, harassment and defamation covers 18% [6]. The serious issues in battling cyberbullying include: finding obscene and offensive words and sentences when it happens on online stages; and then forwarding these cases in Bangladesh in order to find the people behind such actions in real life. No present online network or internet based social sites (for instance, Facebook and Twitter) fuses a framework to naturally and brilliantly distinguish animosity and occurrences of online provocation on its foundation. Due to the non-reality of this significant issue prior, it isn't viewed as the issue of exploration, yet now it is in a dangerous stage. Nobody can overlook this impact on the digital stage which is why it has become an important part of research on how to deal with this issue efficiently. It requires a genuine consideration by analysts and cybercrime agencies to control this movement on online harassment [5]. Ergo, detecting words and sentences which is considered as online harassment on social networking platforms and the extent of the offensive word used in Bengali language is the goal of this work.

## II. LITERATURE REVIEW

Significant amount of work has been done in the field of Natural Language Processing introducing diverse techniques to handle text data. Polarity of text data was calculated in some literature [7, 27]. They divided sentiment analysis into three parts accordingly: document level, sentence level and finally the entity and aspect level. After performing data cleaning and preprocessing and stemming, a score was calculated based on the positive and negative dictionary. The ultimate result allows us to know whether the sentence given as input was positive, negative or neutral. They successfully managed to assign positive, negative or zero values to the sentences being provided as input leading to an ultimate review of the article. Improved baseline algorithm for sentiment analysis was proposed based on the focus sentence and context [8]. They were able to discover what to focus particularly in a given sentence and how to deal with the dynamic sentiment of the word. Another study proposed an efficient algorithm using the methods of NLP and Machine learning for analyzing the social comment and identified whether it was aggressive or not [9]. An effective classifier acts as the core component in their final prototype system that helps to detect cyberbullying on social media. Logistic Regression and Random Forest Classifier trained on the feature stack performed better than Support Vector Machine and Gradient Boosting Machine in this particular case. A large number of abusive words and insults get missed out from the vocabulary because of the vast usage in many different forms and in different languages. Two new speculations for feature extraction were introduced in literature which can be useful in distinguishing cyberbullying [10]. They assembled a model which anticipated remarks as bully or non-bully. The steps involved are normalization, standard feature extraction, additional feature extraction, feature selection and finally classification. In standard feature extraction, they use N-gram, counting and TF-IDF score to construct feature vectors. The ultimate product is the likelihood of the remark being hostile to members. Results show that their speculation expands the precision by 4% and can be utilized to distinguish the remarks that are focused towards peers.

The accuracy of sentiment classification from text was improved effectively compared with traditional CNN and confirms the effectiveness of sentiment analysis based on CNNs and SVM [11]. Emotion classification is divided into supervised, unsupervised and semi supervised methods. Pre-trained word vector was used as input, CNN was used as an automatic feature learner and SVM was used as an automatic emotional classifier. Accuracy rate of using CNN-SVM model combined was found much higher than that of SVM and CNN separately. Another work [12] tried to build lexicon-based word vectors to predict text sentiment. Word embedding was used to build sentiment lexicon, then the polarity of sentiment words from a dataset of user comments is judged and finally naive Bayes was used to classify the represented features on massive dataset and user reviews from the app store. The precision and recall rate of lexicon built by word2vec was larger and easier than the lexicon built by PMI. The average recall was very low as many of the sentiment words in the comments were not in the lexicon.

A recent work used two different types of dataset to detect hate speech using RNN to figure out which model performs better [14]. Dataset-A has little amount of data, Dataset-B's data is more than three times that of Dataset-A. From the general point of view, utilizing Dataset-B can show signs of improvement in execution. As per their results, the good performance was obtained using when the dataset was small. And the good results can be obtained by using deep learning when more data were used for their experiments. They also extracted some data from the raw data at the ratio of 10%-90% as test data and found that the overall performance of Logistic Regression was better than SVM. In the case of using SVM, the performance of TF-IDF was more prominent. From the most recent exploratory advancement, it may be presumed that utilizing BiRNN can improve results. Sentiment analysis using a combination of Naïve Bayes and a lexicon-based algorithm was used to analyze the opinion of different traders and predict the overall sentiment in foreign exchange markets [19]. They managed to achieve a 90% accuracy based on the results. Classification of tweets into positive, negative, and neutral on views of a particular product was done in another study [21]. TextBlob was used to process textual data and Naïve Bayes was used for classifying the text, which is based on Stanford Natural Language Toolkit (NLTK). The training dataset was given to a feed-forward neural network and output layer determined the overall polarity. Then a confusion matrix calculates the accuracy of the results, which was found to be 79-87%. After collecting data from twitter using GloVe embedding's, a study was conducted to remove unwanted URLs, tags, and stop words [22]. Later on, they did POS tagging to label the words. Afterward, the MSP model was used for aspect-based sentiment analysis and the probability was checked based on the polarities. Based on the results accuracy obtained was 74.66%, which proves that the MSP model increases the accuracy in contrast to other neural network models that were used previously. A systematic study concentrated on applying sentiment analysis in a mixed dataset of multiple languages [20]. They used a Bengali-English dataset and a Telugu movie review dataset, which were passed through a single layer CNN, for classification. Using pooling and dropout regularization they managed a 73.2% accuracy in the mixed Bangla dataset and 51.3% in the Telugu movie review dataset. The accuracy could have been optimized if they used Word2vec instead of word indexing.

With the growth of usage of Bangla language online, Bangla text processing has become an emerging field of study. Though Bangla text data is not abundant to analyze, many studies have been conducted on processing Bangla text. A superior way for Bangla sentiment analysis was introduced where they used three different models and two of them are using Word2vec models and another is the traditional Word to Index base text classifier model [15]. Then skip-gram and CBOW were used to generate vector representation of words to feed into a Deep Long Short-Term Memory (LSTM) network. Through their works, they managed to get 83.79% accuracy. Another work aimed to present a Bangla corpus specifically targeted for sentiment analysis [16]. They stemmed their word list using two different stemmers, StemmerR and StemmerP, and generated a word cloud based on the polarity of the words. They managed to successfully polarize the words and achieve a good accuracy based on the sentiment of the words. A study was conducted to determine positive or negative sentiment for Bangla language with higher accuracy and a simpler model [17]. They used a random forest classifier after POS tagging the words and handling negation words. They managed to get 87% accuracy but failed to deal with emoticons and certain characters that sometimes express sentiments too. Bangla sentences were represented based on characters and information were extracted from the characters in another study [18]. They used an embedding layer with 67 units, 3 hidden layers where two layers are with 128 GRU units each and one vanilla layer with 1024 units stacked up serially, and at last, the output layer [28]. They managed to get good accuracy but the usage was limited due to the lack of application. A method for producing Bangla word clusters based on semantic and contextual similarity was proposed in literature [24]. They proposed an unsupervised machine learning technique to develop Bangla word clusters based on their semantic and contextual similarity using n-gram language model. They used tri-gram model equations for calculating similarity. Another study has introduced a word prediction model that can suggest the most probable word in the sentence [23]. They collected their sample data from several lists of newspapers, blogs, social networking sites (i.e. Facebook) and from an open-source Bangla text corpus. Then a training corpus was built from the data in a particular format. A probabilistic algorithm was used to generate the most likely candidates as a suggestion. A semantic similarity checking algorithm was used to omit inappropriate candidate words from the suggestion list. To measure the semantic similarity of a sentence after predicting a word, they used word2vector model and the Stupid Backoff model was used to detect the most probable word. The model emphasizes on scores rather than the probabilities and works well for large n-grams. Based on the results they managed to get an 83.01% accuracy which outperforms other words prediction models that were done before.

### III. WORKFLOW

The workflow of the study is shown in Fig 1. After collecting the dataset maintaining proper standards, we preprocessed the data. Then, we have calculated vectors representing each word in multidimensional space, known as Word embedding, and fed them into hybrid neural network model with various variations. At last, we have analyzed the performance and came to conclusion.

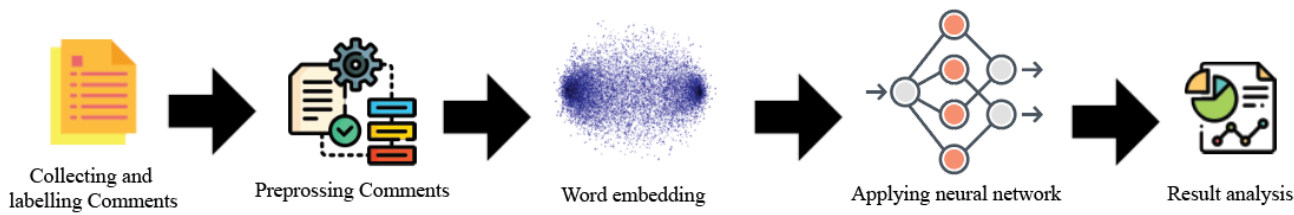

Fig. 1 Workflow of the work

### IV. DATASET AND PREPROCESSING

The dataset that has been used in this work contains comments from the interaction section under posts given by actors, social media influencers, singers, politicians, sportsmen that can be viewed publicly on the Facebook platform [29]. The total number of comments collected is 44001. According to our dataset, 31.9% comments are targeted towards male victims and 68.1% comments are aimed towards female victims. Furthermore, 21.31% comments are targeted towards victims who are social influencers, 5.98% comments are targeted towards politicians, 4.68% comments are targeted towards athletes, 6.78% comments are targeted towards singers and 61.25% comments are targeted towards actors. Table 1 describes the variables that has been used in the dataset. The labels of the dataset are explained below.

*Non-bully:* In general, non-bully comments are those that are not intended to personally attack a person. For example: "তুমি এগিয়ে চলো, আমরা আছি তোমার সাথে।".

*Sexual:* The comments which spread hatred sexually against a person, sexually harass a person are categorized into this class. For example: "এভাবে অর্ধনগ্ন ছবি পোস্ট না করে, পর্ণগ্রাফি করলেই পারেন।".

*Threat:* The threat class consists of comments input by users making a threat against another person to harm or kill. For example: "তোকে জুতাপিটা করে দেশ থেকে বিতাড়িত করা উচিত।"

*Troll:* The comments given by users to mock a person hurtfully are labelled with this category. For example: "এই অপদার্থ মহিলার চেহারা দেখলেই যেন বমি আসে।".

*Religious:* The comments, which state offensive words or convey and spread hatred against any religious groups, are categorized into this class. For example: "এদের দেখতে মানুষের মত লাগে, কিন্তু এরা আসলে মূর্তিপূজারি কাফির।"

Table 1 Variables in Dataset

| Variable | Type | Description |
| --- | --- | --- |
| Comment | Text | Collected comments of the users |
| Category | Categorical | Occupation of the victim |
| Gender | Categorical | Gender of the victim |
| Number of React | Integer | Number of likes or reactions on that comment |
| Label | Categorical | Type of harassment |

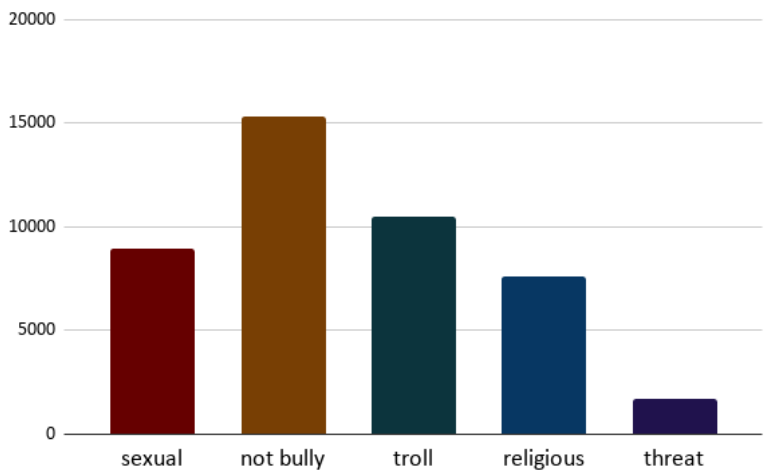

Fig. 2 Number of comments in each category

### A. Data Preprocessing

After collecting the comments, we removed bad characters, punctuations, etc. from the raw data and pre-processed the collected information in order to feed it to our neural network. We performed the pre-processing steps in three parts: Stop words Removal, Tokenization of String and Padded sequence conversion.

Stop words [26] are frequently used prepositions, linkers, quantifiers etc in any natural language. 'The', 'is', 'in' etc are the most common stopwords in English language. In Bengali language, we have similar kind of stopwords as well, such as - 'অতএব', 'অথচ', 'অথবা' etc. We found a list of Bengali stopwords [13] and removed 398 stopwords from our raw sentences. It helps to improve accuracy and efficiency as we only work with meaningful words and vocabulary size gets increased. It also filters out most of the spam.

We used a property of tensorflow, which is called 'Tokenizer' in order to put a value to the most frequently used words. This will create a dictionary with the key being the word and the value being the token for that particular word. In the next step, we tried to turn the sentence into a list of values based on these tokens. We also replaced the unseen and unknown words, which don't exist in the word index with "OOV". Words like 'নাস্তিক' , 'বিশ্বাস' , 'আল্লাহ' etc. came up to the top as these words appeared most frequently in our dataset. After tokenizing, we got a large number of tokens, which consists of 72202 vocabularies.

While feeding data into the neural network for training, we need them to be uniform in size. We used padding to convert the sentences into a uniform sized text sequence. Once the tokenizer has created the sequences, these sequences are passed to pad sequences in order to get padded in similar shape. As a result, the list of sentences has padded out into a matrix and each row of the matrix has a max length of 120. For smaller sentences, we had to put an appropriate number of zeros after the sentence.

### B. Word Embedding

`To represent each of the words, we have used word-embedding vectors for each token of the sentence. Word embedding is a process where words and associated words are clustered together in a multi-dimensional vector space. We have words in a sentence and often words that have similar meanings are close to each other. Similar vectors are given to the words those were found together in a higher dimensional space. Therefore, words can begin to cluster together. The meaning of the words came from the labelling from our dataset. We used the Word2Vec [25] embedding model. We considered 19469 vocabularies and embedding dimension was set to 16 in our model. As a result, we can visualize our words clustering mainly on two opposite sides. A visualization of the embedding vectors in multidimensional space is shown in Fig 3.

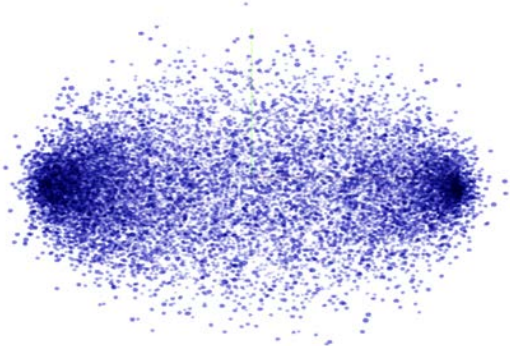

Fig. 3 Visualization of word embedding

**V. METHODOLOGY**

Our proposed model can be divided into three parts. At first, we considered these labels - 'sexual', 'threat', 'troll', 'religious' as 'bully' and applied a binary classification model to identify whether the comment is harassment or not. Then, we developed a hybrid model to classify all 5 classes. Finally, we collected the predicted result from both binary and multi-class classification models and applied ensemble method in order to improve our accuracy. The setup is shown in Fig 4.

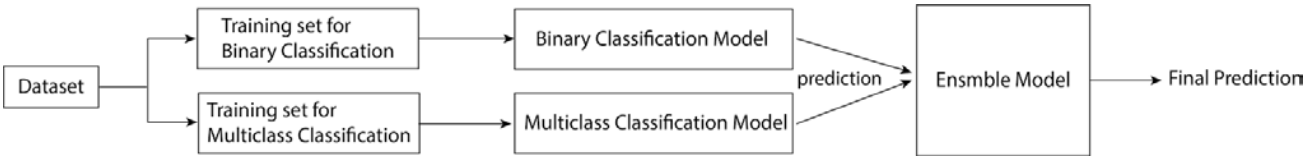

Fig. 4 Experimentation Setup

### A. Binary Classification

Objective of this classification was to predict if a comment is labelled as 'bully' or 'not bully'. Our implemented binary classification neural model is depicted in Fig 5. After embedding the raw inputs, we applied a 1D convolution layer. We set the number of output filters as 32, length of the 1D convolution window as 3 and used 'relu' as the activation function. Now, words will be grouped into the size of 3 and convolutions will be learned that can map classification to the desired output. Next, we implemented a LSTM layer for faster training and good performance. In this layer, we declared the number of outputs as 100. To avoid overfitting, we used dropout and recurrent dropout with a rate of 0.2. In the next layer, we used global average pooling 1D, which averages across the vector to flatten it out. Afterwards, the outputs were fed into a shallow NN. In this step, we use a dense layer with a 'relu' activation function and another dense layer with a 'sigmoid' activation function. Finally, we compiled the neural network with 'binary_crossentropy' as we are classifying two different classes only.

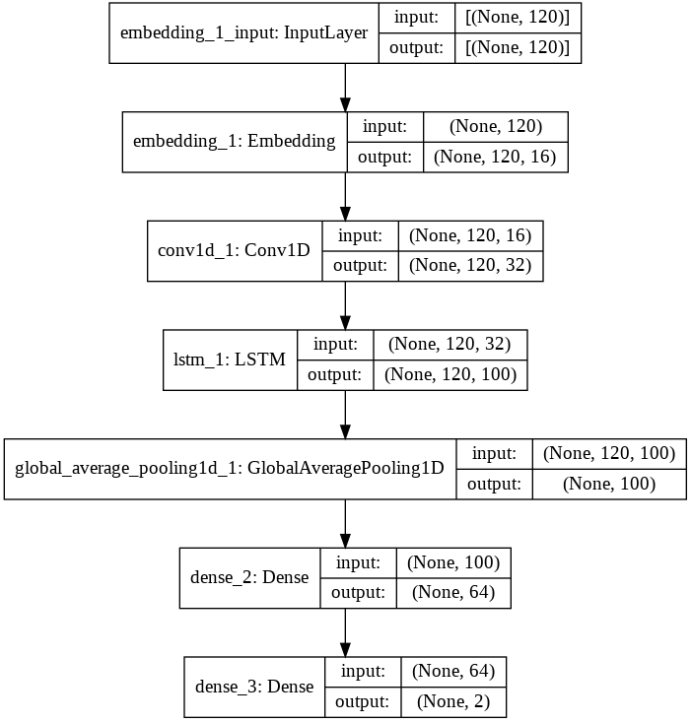

Fig 5 Binary Classification Model

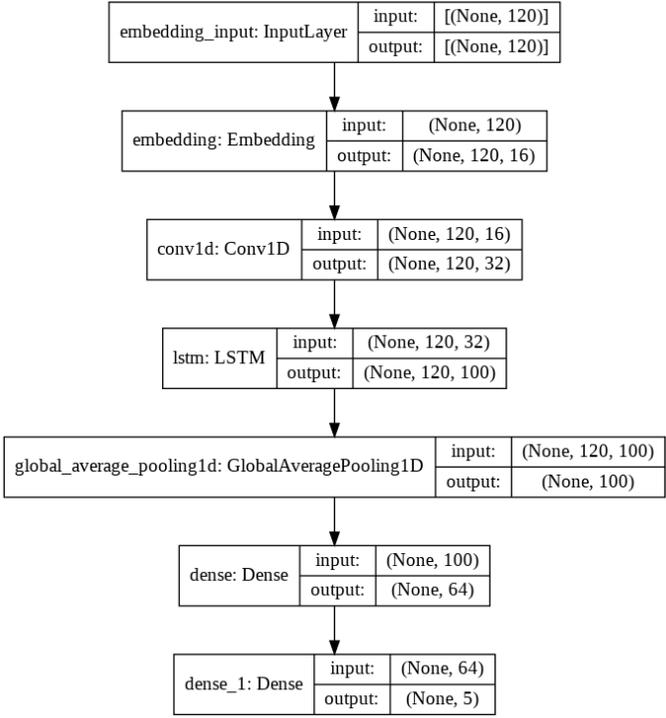

Fig 6 Multiclass Classification Model

## B. Multiclass Classification

Objective of this classification is to predict whether a comment is a bully or what kind of bully. Therefore, there are five classes - 'not bully', 'sexual', 'troll', 'religious', and 'threat'. In this model, we have used the similar kind of structure of binary classification. However, we have used 'softmax' activation function in the last layer of the DNN to predict the probability of each example to belong into these classes. Fig. 6 shows the architecture of this model.

VI. EXPERIMENTAL RESULT

## A. Binary Classification

In Binary Classification, we have used 15 epochs and each epoch took an average of 264.8 seconds. The best validation loss we got was 0.27204. Furthermore, the validation accuracy of this model was 87.91%. This classifier model holds precision of 90%, recall of 75% and F1-score of 82%. This model can successfully predict 95% of the 'not bully' comments and 75% 'bully' comments. Epoch Vs Accuracy graph for Binary classification model is shown in Fig 7.

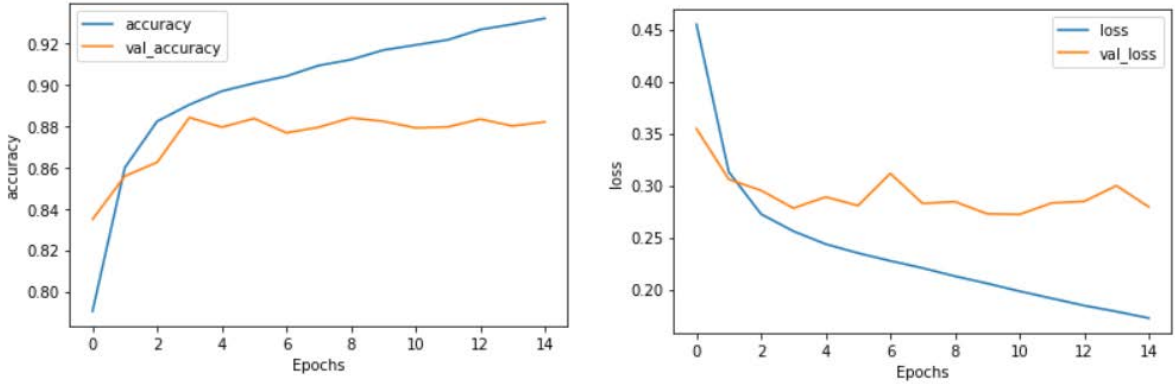

Fig. 5 Epoch vs Accuracy and loss for binary classification

## B. Multiclass Classification

In Multiclass Classification, we have used 15 epochs and each epoch took an average of 261.4 seconds. The best validation loss we got was 0.6210. Furthermore, the validation accuracy of this model was 79.29%. This classifier model holds precision of 81%, recall of 74% and F1-score of 76%. This model can successfully predict 85% of the not bully comments and 82% religious, 80% sexual, 48% threat and 73% troll comments. Epoch Vs Accuracy graph for Multiclass classification model is shown in Fig 8.

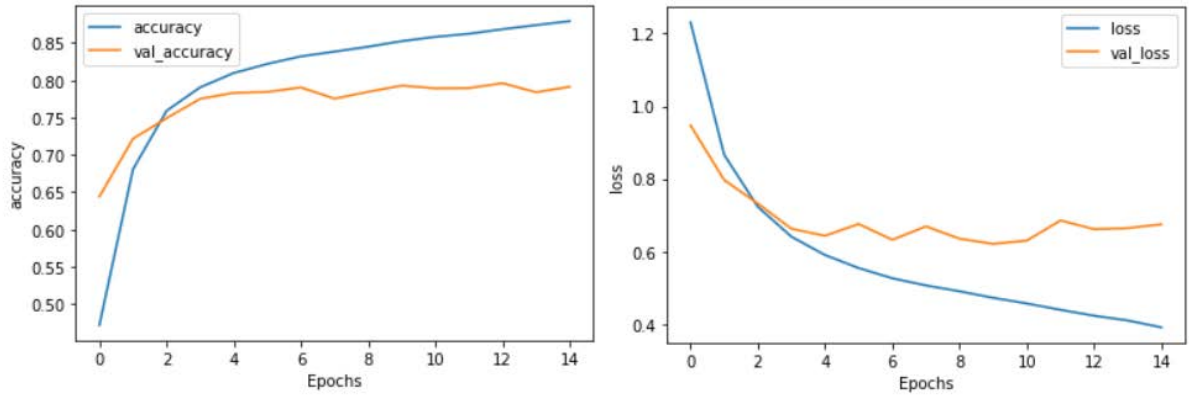

Fig. 6 Epoch vs Accuracy and Loss for multiclass classification

## C. Ensemble

To improve the accuracy of multiclass classifier, we tried to use ensemble model techniques with the help of binary classifier. We trained using the predicted result for all the comments from both of the models and tried different supervised machine learning algorithms, such as: Random Forest, SVM, KNN, Naïve Bayes etc. classifiers. SVM algorithm stood out among these algorithms with an improved accuracy of 85%. Performance comparison of these algorithms is shown in Table 2. According to the confusion from categorical classification model matrix shown in Fig 9, we could successfully predict 91% of 'Not bully' comments, 85% of 'Religious' comments, 81% of 'Sexual' comments 50% of 'Threat' and 84% of 'Troll' comments. Detailed information of precision, recall and f1-score of each class using SVM are shown in Table 3.

Table 2 Performance of different algorithm in Ensemble Technique

| Algorithm | Accuracy | Precision | Recall | F1 score |
|---|---|---|---|---|
| Random Forest | 0.84 | 0.84 | 0.84 | 0.84 |
| **SVM** | **0.85** | **0.85** | **0.85** | **0.84** |
| KNN | 0.84 | 0.85 | 0.84 | 0.84 |
| Naïve Bayes | 0.79 | 0.78 | 0.79 | 0.78 |

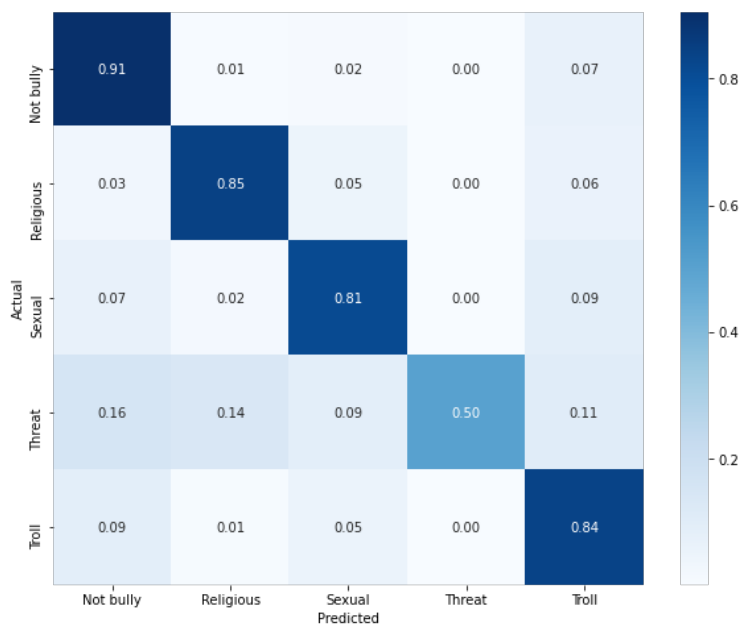

Fig. 7 Confusion Matrix of Final Prediction

Table 3 Performance of different algorithm in Ensemble Technique

| Class | Precision | Recall | F1 score |
|---|---|---|---|
| Not Bully | 0.87 | 0.91 | 0.89 |
| Religious | 0.91 | 0.85 | 0.88 |
| Sexual | 0.84 | 0.81 | 0.83 |
| Threat | 0.90 | 0.50 | 0.65 |
| Troll | 0.77 | 0.84 | 0.80 |

## VI. RESULT ANALYSIS

Our proposed model can classify the non-bully sentence as well as different category of bully successfully. In case of binary classification, it gave 87.91% accuracy, which is better than the recent studies discussed in the literature. Further, we built a multiclass classification model to identify all the category of harassment and got 79.29% accuracy. To improve this accuracy with the help of binary classification we used an ensemble technique and got 85% accuracy for classifying into different category of harassment.

We managed to get a fair accuracy comparing with the recent work in Bangla language processing discussed in the literature review. However, in the multiclass classification, the 'threat' category showed comparatively less accuracy than other categories due to low recall value. But the precision is very high in this category. So, it doesn't falsely classify any other comments as threat. Again, low recall shows that it cannot effectively identify all the threat comments due to the insufficient training data on 'threat' category. It can be improved in the future if more data is given to the training set.

Another finding is that it showed false positive results for comparatively longer and complex sentences. For example: "জুতা তৈরির ফ্যাক্টরীর পাশেই পশু জবাই করতে দেখা যায়". This is a sentence that is not a bully or harassing sentence. However, if we enter this sentence as an input, the result identifies it as a threat (95%). Since, "জুতা" and "জবাই" - these words are often used to express verbal threat towards someone in Bengali language. Most of the sentences that contain words like these were labelled as a threat in our dataset. As our dataset was only based on social media comments, we had a limitation and had to work with the comments that we were able to collect. Nevertheless, this issue can be solved, if a mass variety of sentences with different word order and syntax are included as a training set.

## VI. CONCLUSION

Bully detection on the Facebook platform for various major languages is quite challenging. This is because of the diversity of different languages and the ways they are used or typed by different users. In this paper, we have focused on detecting bully expressions on the Facebook platform for Bengali language. We can successfully detect whether a statement input by a user is a bully expression or not. Besides bully detection, we can also state to what category it falls in and to what extent it falls. We have made five labels: non-bully, sexual, threat, troll and religious to categorize what sort of a statement has been input. The dataset used for this paper is up to date consisting of a decent amount of data enabling us to get high precision results. Our model took a higher training time and sometimes showed a false positive result for comparatively long sentences. The drawbacks of this paper can be improved for future endeavors.